\documentclass[letterpaper, 10 pt, conference]{ieeeconf}  

\IEEEoverridecommandlockouts    

\usepackage{graphicx}
\usepackage{cite}
\usepackage{color}
\usepackage{amsmath,amssymb} 
\usepackage{epsfig} 
\usepackage{multirow}
\usepackage{hyperref}
\usepackage{bm}
\usepackage{caption}
\usepackage{copyrightbox}
\usepackage{subcaption}

\usepackage{soul,xcolor} 

\title{\LARGE \bf
Maneuver-Aware Pooling for Vehicle Trajectory Prediction
}

\author{Mohamed Hasan$^{1}$,  Albert Solernou$^{2}$, Evangelos Paschalidis$^{2}$, He Wang$^{1}$,\\ Gustav Markkula$^{2}$  and Richard Romano$^{2}$
\thanks{$^{1}$ School of Computing, University of Leeds}%
\thanks{$^{2}$ Institute for Transport Studies, University of Leeds}%
}

\newcommand{\quotes}[1]{``#1''}

\begin{document}

\maketitle
\thispagestyle{empty}
\pagestyle{empty}

\begin{abstract}

Autonomous vehicles should be able to predict the future states of its environment and respond appropriately. Specifically, predicting the behavior of surrounding human drivers is vital for such platforms to share the same road with humans. Behavior of each of the surrounding vehicles is governed by the motion of its neighbor vehicles. This paper focuses on predicting the behavior of the surrounding vehicles of an autonomous vehicle on highways. We are motivated by improving the prediction accuracy when a surrounding vehicle performs lane change and highway merging maneuvers. We propose a novel pooling strategy to capture the inter-dependencies between the neighbor vehicles. Depending solely on Euclidean trajectory representation, the existing pooling strategies do not model the context information of the maneuvers intended by a surrounding vehicle. In contrast, our pooling mechanism employs polar trajectory representation, vehicles orientation and radial velocity. This results in an implicitly maneuver-aware pooling operation. We incorporated the proposed pooling mechanism into a generative encoder-decoder model, and evaluated our method on the public NGSIM dataset. The results of maneuver-based trajectory predictions demonstrate the effectiveness of the proposed method compared with the state-of-the-art approaches. Our \quotes{Pooling Toolbox} code is available at https://github.com/m-hasan-n/pooling.



\end{abstract}

\section{INTRODUCTION}

Human drivers need to predict the intentions of other road users to safely drive vehicles and navigate through other traffic. This is critical when making decisions like yielding to a coming vehicle or merging into traffic \cite{chai2019multipath}. Similarly, to be fully autonomous, an Autonomous Vehicle (AV) should be able to predict the future states of its environment and respond appropriately. This is crucial for the planning and control of such self-driving platforms. Specifically, predicting the behavior of human drivers becomes necessary for AVs to share the same road with humans \cite{gupta2018social}. 

Human driver behavior on a highway is highly influenced by the other neighboring drivers. While experienced human drivers are able to confidently predict future behaviors of other drivers, it is still an open problem for AVs due to the inherent challenges of maneuver prediction. Some of these challenges can be arising from: (1) trajectories tend to be non-linear over long time horizons, (2) the inherent uncertainty about latent variables such as drivers' motivations and goals, and (3) the intrinsic multi-modality nature of the decision-making processes, reflected in the multiple possible decisions under the same traffic condition \cite{deo2018convolutional}.

We address the problem of trajectory prediction of the human-driven vehicles surrounding an autonomous vehicle on highways. In line with the existing work \cite{alahi2016social, deo2018convolutional}, this is achieved by iteratively predicting the trajectory of each surrounding vehicle around the AV. We call the subject surrounding vehicle to be predicted as \quotes{\textit{ego}} and the vehicles around the ego vehicle as \quotes{\textit{neighbors}}.  Specifically, given the observed motion trajectories of human drivers for the immediate past (e.g. 3 seconds) of an ego vehicle and all its neighbor vehicles, predict the future trajectory of the ego vehicle considering all its neighboring vehicles.

This problem has been tackled in the literature (for both vehicle and pedestrian agents) from two broad aspects. First, encoding the status of the neighbors by a pooling vector \cite{alahi2016social, deo2018convolutional, gupta2018social} that captures their inter-dependencies. Second, using generative models that output multi-modal probability distribution over the future trajectory of the ego agent \cite{chai2019multipath, deo2018convolutional, gupta2018social, hu2019multi, zyner2019naturalistic}. 

We focus on the first aspect and propose a novel pooling mechanism to capture the interaction between the neighbor vehicles. Previous pooling strategies represent vehicle trajectories using longitudinal and lateral positions. It is reasonable to assume that human drivers employ higher-order information to reason over the surrounding traffic. Thus, We hypothesize that the prediction accuracy can be improved by incorporating higher-order information like vehicles orientation and velocity in the pooling operation. 


We are motivated by improving the motion prediction accuracy when vehicles perform lane change and highway merging maneuvers. Thus, we introduce a pooling strategy that is based on signals from the orientation and radial velocity of the vehicles. Hence, the resulting pooling vector is implicitly aware of the maneuvers intended by the surrounding vehicles.

We incorporated the proposed pooling strategy into a generative encoder-decoder model, that outputs a multi-modal distribution of predictions conditioned on a set of semantic maneuvers. The overall model is evaluated on the public NGSIM dataset \cite{colyar2007us, colyar2007i80} and the results are compared to state-of-the-art pooling approaches. In addition to the traditional overall evaluation of the prediction accuracy, we perform a finer, maneuver-based evaluation where the accuracy is also reported for lane change and highway merging maneuvers.  
 
The main contribution of our work is a maneuver-aware pooling mechanism that outperforms the current pooling approaches in terms of prediction accuracy as shown over an extensive maneuver-based evaluation. To help reproduce the proposed and other \cite{alahi2016social, deo2018convolutional, gupta2018social} pooling approaches, our \quotes{Pooling Toolbox} code is publicly available.

\begin{figure*}
\vspace{2mm}
\centering
\begin{subfigure}{0.3\textwidth}
    \includegraphics[width=\textwidth]{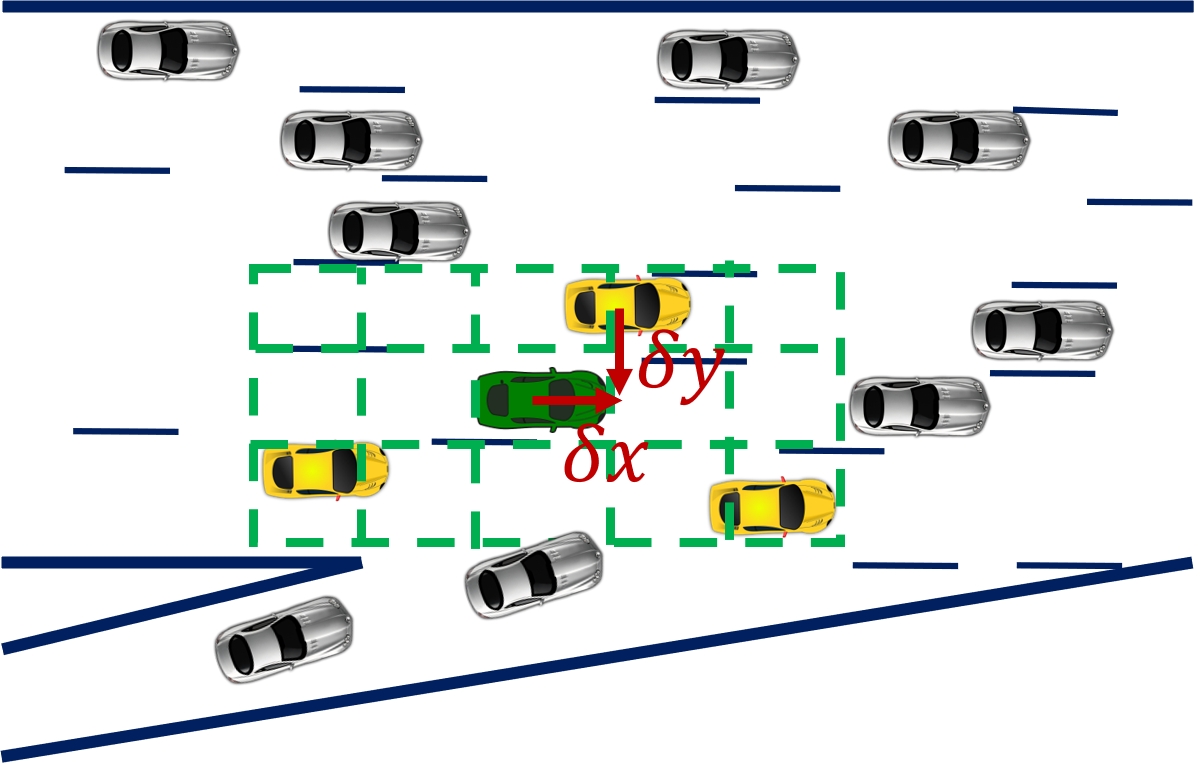}
    \label{subfig:social}
\end{subfigure}
\hfill
\begin{subfigure}{0.3\textwidth}
    \includegraphics[width=\textwidth]{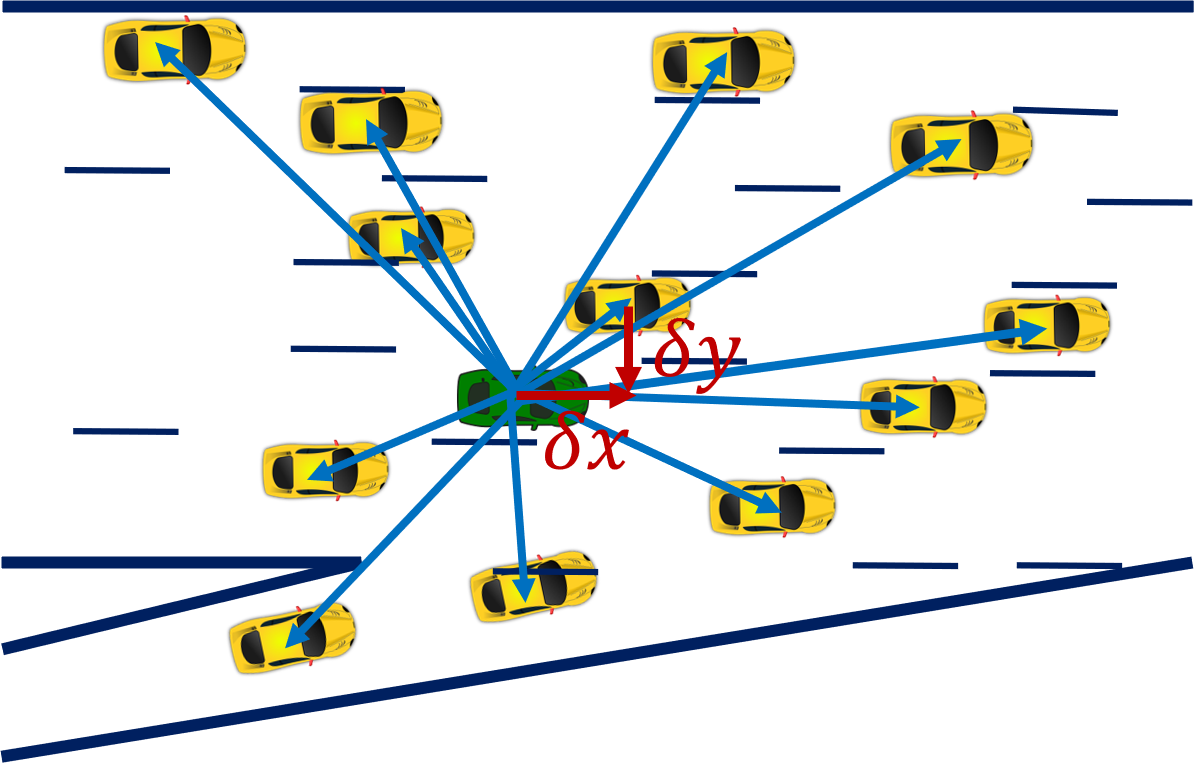}
     \label{subfig:sgan}
\end{subfigure}
\hfill
\begin{subfigure}{0.3\textwidth}
    \includegraphics[width=\textwidth]{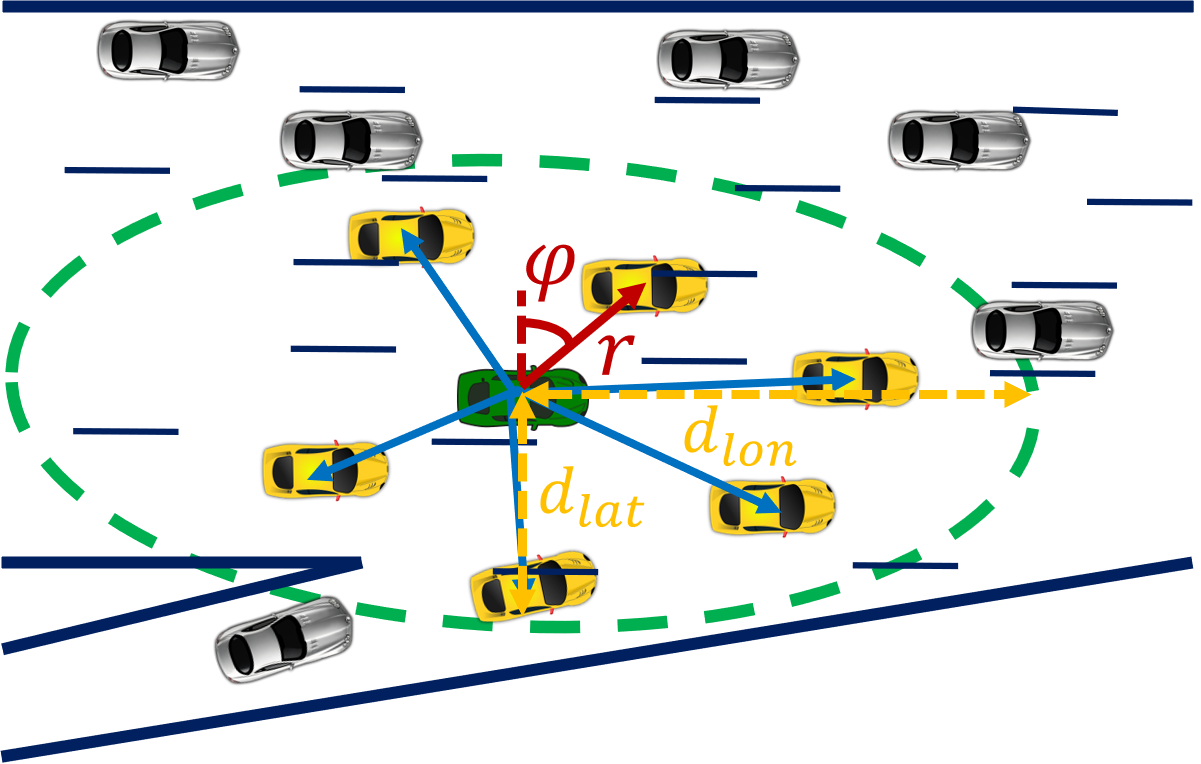}
     \label{subfig:polar}
\end{subfigure}     
\caption{Visualizing pooling mechanisms (A green vehicle shows the ego, yellow vehicle shows a neighbor covered by the pooling strategy, and grey vehicle shows a non-covered neighbor). Left: a spatial grid is centered around the ego vehicle. The social tensor is structured accordingly and populated with LSTM states of the ego and exisiting neighbor vehicles \cite{alahi2016social, deo2018convolutional}. Center: relative positions between the ego vehicle and all its neighbors are concatenated to vehicle LSTM states \cite{gupta2018social}. Right: the proposed pooling strategy where vehicle LSTM states are concatenated to relative polar positions (distance and angle) rather than the Cartesian representation used by the previous works.}
\label{fig:pooling}
\end{figure*}

\section{Related Work and Background}

\subsection{Generative Encoder-Decoder Models}
Recurrent Neural Networks (RNNs) represent a rich class of dynamic models that extend feedforward networks for sequence generation in diverse domains like speech recognition \cite{chorowski2014end}, machine translation \cite{chung2015recurrent} and image captioning \cite{vinyals2015show}. Long-Short Term Memory (LSTM) is an instance of RNNs that has been proved successful in sequence learning and generation tasks \cite{graves2013generating, alahi2016social, vinyals2015show}. Since motion prediction can be considered as a sequence generation task, and inspired by the LSTM success in this domain, a number of RNN-based approaches have been proposed for trajectory prediction \cite{alahi2016social, gupta2018social, deo2018convolutional, zyner2019naturalistic, altche2017lstm, park2018sequence, xue2018ss}  of both pedestrians and vehicles. 

In their seminal work Social LSTM, Alahi et al. \cite{alahi2016social} extended RNNs for human trajectory prediction using a social pooling layer that models nearby pedestrians. Gupta et al. proposed a Generative Adversarial Networks \cite{goodfellow2014generative} (GAN): a RNN Encoder-Decoder generator and a RNN based encoder discriminator, to predict socially-acceptable multimodal pedestrian trajectories \cite{gupta2018social}. Social LSTM approach was further improved in \cite{deo2018convolutional} by using convolutional social pooling applied to vehicle motion prediction on highways. LSTM network is also used to predict the location of vehicles in an occupancy grid \cite{kim2017probabilistic} at different future intervals. Convolutional networks and LSTM were combined to predict multi-modal trajectories for an agent on a bird's eye view image \cite{bhattacharyya2018accurate}.

\subsection{Multi-modal Predictive Distribution}

Deo et al. proposed a model \cite{deo2018convolutional} that outputs a multi-modal predictive distribution over future trajectories based on three lateral and two longitudinal maneuver classes. MultiPath model \cite{chai2019multipath} can predict a discrete distribution over a set of future state-sequence anchors and output multi-modal future distributions. The GAN based encoder-decoder architecture in \cite{gupta2018social} encourages diverse multimodal predictions of pedestrian trajectories with the introduced variety loss. A multi-modal probabilistic prediction approach was presented in \cite{hu2019multi} based on a Conditional Variational Autoencoder and is capable of jointly predicting sequential motions of each pair of interacting vehicles. Zyner et al. \cite{zyner2019naturalistic} presented a model based on RNN with a mixture density network output layer, for predicting driver intent at urban single-lane roundabouts through multi-modal trajectory prediction with uncertainty.

\subsection{Pooling Mechanisms}
\label{sec:pooling_review}

Social LSTM (S-LSTM) has been introduced for pedestrian motion prediction \cite{alahi2016social}, where one LSTM is assigned to each person in a scene. To capture the interaction of people in a neighborhood, neighboring LSTMs are connected through a “Social” tensor. This tensor is constructed by defining a spatial grid around the agent being predicted (Fig. \ref{fig:pooling} left), and populating the grid with LSTM states of this agent and its neighbors based on their spatial configuration. A fully connected and a SUM pooling layers are then applied to the social tensor to produce the pooling vector. 

Deo et al. introduced convolutional social pooling (CSP) \cite{deo2018convolutional} to predict vehicle motion behaviors on highways. They extended the work of \cite{alahi2016social} improving the pooling layer by using convolutional and Max-pooling (rather-than a fully-connected and SUM-pooling) layers to embed the LSTM states populated in the social tensor.   

In their Social-GAN (S-GAN) work \cite{gupta2018social}, Gupta et al. argued that the grid-based pooling scheme presented in Social Pooling \cite{alahi2016social, deo2018convolutional} is a hand-crafted solution that is slow and fails to capture global context. They proposed to concatenate the relative position between an agent and \textit{all} its neighbors (Fig. \ref{fig:pooling} center) with the LSTM state of each agent. The concatenated tensor is then processed independently by a Multi-Layer Perceptron and a Max-pooling layer. 

The previous approaches depend solely on the lateral and longitudinal positions to represent the vehicle trajectories. In contrast, we employ polar coordinates, vehicles orientation and radial motion velocity to represent the vehicle trajectories.  
Using such higher-order information results in an implicitly maneuver-aware pooling strategy, that proved to outperform the state-of-art pooling approaches in terms of the prediction accuracy.  

\section{Problem Formulation}

Inline with the previous approaches \cite{alahi2016social, deo2018convolutional, gupta2018social, chai2019multipath}, We formulate the problem of vehicle trajectory prediction as estimating the probability distribution of the possible future trajectories of an ego vehicle conditioned on its track history and track histories of the neighbor vehicles around it, at each discrete time step \(t\). Given observations \(X\) in the form of past trajectories of the ego and its neighbor vehicles, our model seeks to provide a distribution over future trajectories of the ego vehicle \(P(Y|X)\).

\subsection{Inputs and Outputs} 
The inputs to our model are track histories:
\begin{align}
\begin{split}
X = [x^{(t-t_h)}, ..., x^{(t-1)}, x^{(t)}]
\end{split}
\end{align}
where \(t_h\) is a fixed (history) time horizon, and at any time instant \(t\),
\begin{align}
\begin{split}
x^{(t)} = [r_0^{(t)}, \phi_0^{(t)}, V_{r_0}^{(t)}, r_1^{(t)}, \phi_1^{(t)}, V_{r_1}^{(t)}, ..., r_n^{(t)}, \phi_n^{(t)}, V_{r_n}^{(t)}]
\label{eq:traj}
\end{split}
\end{align}
represents the position (represented by a distance \(r\) and an angle \(\phi\)) and radial velocity \(V_r\) of the ego vehicle  (subscript \(0\)), and the neighbor vehicles (subscripts \(1\) to \(n\)), represented in polar coordinates.  

The output of the model is a probability distribution over:
\begin{align}
\begin{split}
Y = [y^{(t+1)}, ..., y^{(t+t_f)}]
\end{split}
\end{align}
where \(t_f\) is a fixed (future) time horizon, and at any time instant \(t\),
\begin{align}
\begin{split}
y^{(t)} = [r_0^{(t)}, \phi_0^{(t)}, V_{r_0}^{(t)}]
\end{split}
\end{align}
represents the future position and velocity of the ego vehicle being predicted.

\subsection{Multi-modal Trajectory Prediction}
\label{sec:multimodal}
The multi-modal distribution over future trajectories can be hierarchically factorized \cite{deo2018convolutional, chai2019multipath}. First, estimate the intent uncertainty over a set of maneuver classes. Second, given the intended maneuvers, predict the vehicle future trajectory.  
 
The intentions of human drivers are modelled using two discrete sets of maneuver types: (1) location-wise maneuvers \(M_l=\{m_{l_p}\}_{p=1}^P\), and (2) acceleration-wise maneuvers \(M_a=\{m_{a_q}\}_{q=1}^Q\). We model the uncertainty over each discrete set of maneuvers using a softmax distribution. 


Given the intended maneuvers \(m\), the uncertainty over the future trajectory is modelled as a Gaussian distribution:
\begin{align}
\begin{split}
P_{\Theta}(Y|m) = N(Y|\mu(X), \Sigma(X))
\end{split}
\end{align}
that is parameterized by \cite{deo2018convolutional}:  
\begin{align}
\begin{split}
\Theta = [\Theta^{(t+1)}, ..., \Theta^{(t+t_f)}]
\end{split}
\end{align}
which are the parameters of a multivariate Gaussian distribution at each time step in the future. At any time  \(t\), this is given by the Gaussian parameters: \( \Theta^t = [\mu^{(t)}, \Sigma^{(t)}]\), corresponding to the mean and variance of the future vehicle position and velocity.

The multi-modal output conditional distribution over future trajectories can now be expanded in terms of the maneuvers:
\begin{align}
\begin{split}
P(Y|X) = \sum P(m|X) P_{\Theta}(Y|m, X)
\end{split}
\end{align}
which yields a Gaussian Mixture Model distribution (GMM) \cite{chai2019multipath}. The mixture weights are defined by the probabilities of the maneuvers.

\section{Maneuver-Aware Pooling}
\label{sec:mnvr_pooling}

The objective of a pooling strategy is to summarize the context of interaction between the ego and surrounding vehicles within a certain neighborhood. While this neighborhood is defined by a spatial grid in Social Pooling \cite{alahi2016social, deo2018convolutional}, it includes all agents in a given scene in the work of S-GAN \cite{gupta2018social}. Similar to S-GAN, we do not restrict the neighborhood to the handcrafted structure of the spatial grid. However, we limit it by lateral \(d_{lat}\) and longitudinal \(d_{lon}\) distances (Fig. \ref{fig:pooling} Right) around the ego vehicle, considered as hyperparameters.   

To be aware of vehicle maneuvers, we represent trajectories in polar form and use information from vehicles orientation and radial motion velocity (\ref{eq:traj}). Relative coordinates are used for translation invariance. The origin \(O\) of a stationary \textit{frame of reference} is fixed at the ego vehicle position at time \(t\). We assume that the sensors and/or computers onboard the AV can measure and/or compute the parameters required for our solution. For example, if the position history of the ego vehicle is measured in terms of lateral (\(x\)) and longitudinal (\(y\)) positions as \( [x_0^{(t-t_h)}, y_0^{(t-t_h)}, ..., x_0^{(t)}, y_0^{(t)}] \), the resulting polar-representation is given by \( [r_0^{(t-t_h)}, \phi_0^{(t-t_h)}, ..., r_0^{(t)}, \phi_0^{(t)}] \), where at any time step \(j\):

\begin{equation}
\begin{split}
r_0^{(j)} &= \sqrt{ {(x_0^{(j)}-x_0^{(t)})}^2 + {(y_0^{(j)}-y_0^{(t)})}^2}\\
\phi_0^{(j)} &= \arctan(\frac{y_0^{(j)}-y_0^{(t)}}{x_0^{(j)}-x_0^{(t)}})
\label{eq:polar}
\end{split}   
\end{equation}
Accordingly, the angle \(\phi_0^{(j)}\) defines the orientation of the ego vehicle at time \(j\). The same representation is also used to compute the relative position between each neighbor and the ego vehicle.


\begin{figure*}
\vspace{3mm}
\centering 
\includegraphics[width=0.7\textwidth]{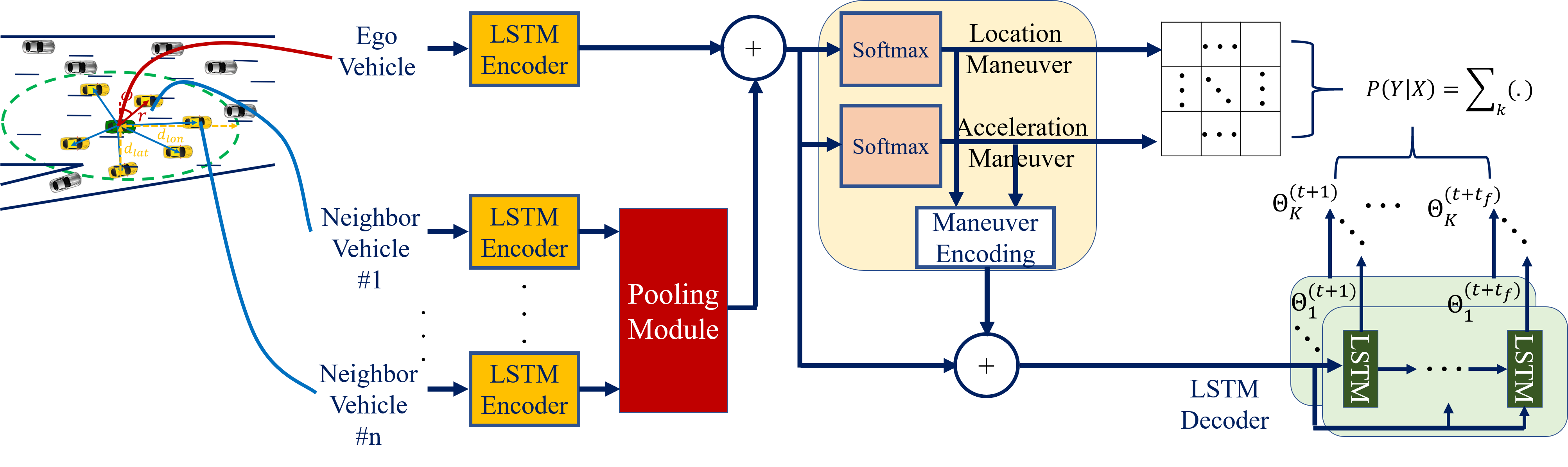}
\caption{LSTM encoder-decoder based model consisting of encoding, pooling, maneuver recognition and decoding modules.}
\label{fig:model}
\vspace{-2mm}
\end{figure*}

We augment vehicle positions by radial motion velocity to better represent the trajectories. Computing radial velocity is based on the vehicle velocity and orientation, which are important signals to provide information of the intended maneuver. Specifically, the radial velocity of the $i$-th neighbor vehicle at time \(j\) is defined relative to the ego as:

\begin{equation}
\begin{split}
V_{r_i}^j = V_i^j \cos{(\theta_i^j - \phi_i^j)}
\end{split}
\end{equation}
where \(\phi_i^j\) is defined similarly to  (\ref{eq:polar}), \(V_i^j\) denotes the linear velocity of the neighbor vehicle relative to the ego linear velocity \(V_0^t\) at the the origin \(O\), and the angle \(\theta_i^j\) defines the orientation of the neighbor vehicle.



\section{Model}

The proposed pooling mechanism is incorporated in the model shown in Fig. \ref{fig:model} consisting of an LSTM encoder, a pooling module, a maneuver recognition module and an LSTM decoder.

\subsection{Encoder}
We encode the state of motion of each vehicle using an LSTM encoder. At any time \(t\), a sequence of \(t_h\) time steps of the track history is passed through the encoder. The LSTM states for each vehicle are updated frame by frame over the \(t_h\) past frames \cite{deo2018convolutional}. The LSTM weights are shared across the sequences of all vehicles \cite{alahi2016social}.   

\subsection{Pooling Module}
\label{sec:pooling}
The interaction of vehicles in a given scene is captured by the pooling module. This is achieved by pooling the LSTM states of the neighbor vehicles around the ego vehicle. The output is a pooling vector summarizing the information needed by the ego vehicle to make a maneuver decision. We extend the pooling mechanism in \cite{gupta2018social} to implement our pooling strategy detailed in Sec. \ref{sec:mnvr_pooling}.  

First, at any time step \(t\) the neighbor vehicles around the ego vehicle are identified. The relative position and radial velocity of the ego vehicle and between each neighbor and the ego are then computed (relative to the position and velocity at the origin \(O\). Second, the relative position and velocity are concatenated with each vehicle’s LSTM hidden state, and processed independently by a Multi-Layer Perceptron (MLP). Finally, the MLP output is Max-pooled element-wise to compute the pooling vector of the ego vehicle. This method can capture the inter-dependencies of the motion of neighbor vehicles, without being restricted to a specific grid size. More importantly, being based on vehicle orientation and radial motion velocity, the pooling operation becomes more aware of the intended maneuvers. 

\subsection{Maneuver Recognition}
\label{sec:mnvr_rec}

The location-wise \(M_l\) maneuvers (Sec. \ref{sec:multimodal}) include three classes: lane keeping, left lane-change and right lane-change, which correspond to semantic location-change concepts on highways. Similarly, the acceleration-wise \(M_a\) maneuvers include: constant speed, speeding up and slowing down classes. A vehicle is annotated with one of the lane-change classes if it changes its lane during the prediction horizon, while marked under the lane-keeping class otherwise. The NGSIM dataset provides the acceleration of each vehicle. A vehicle is annotated with a slowing/speeding class if its mean future acceleration is less/larger than \(-a\)/\(+a\) threshold, while marked under the constant-speed class otherwise.   

The maneuver recognition module consists of two softmax layers to recognize the location and acceleration classes. The input to this module is the LSTM state of the ego vehicle augmented by the pooling vector from the pooling module. Each softmax layer outputs the probability of each maneuver class. The maneuver encoding block in Fig. \ref{fig:model} encodes the output from each softmax layer into a one-hot vector. Both one-hot vectors are concatenated with the trajectory encoding and the resulting tensor is passed to the decoder module.

\subsection{Decoder}
An LSTM decoder is used to generate the conditional distribution over future trajectories of the ego vehicle. At any time \(t\), The decoder generates the future trajectories over the next \(t_f\) time steps. At each time step, the decoder outputs a multi-modal distribution conditioned on the maneuver classes.    

\section{Experiments and Results}

\subsection{NGSIM Dataset}

The NGSIM public dataset is used for our experiments. This dataset consists of two subsets: US-101 \cite{colyar2007us} and I-80 \cite{colyar2007i80} of trajectories of real freeway traffic captured at 10 Hz. Each subset consists of three 15-min segments recorded over a time span of 45 minutes. The recorded segments represent mild, moderate and congested traffic conditions. 

The dataset provides the longitudinal and lateral co-ordinates of vehicles projected to a local co-ordinate system, in addition to other data like velocity and acceleration. We divide the complete NGSIM dataset into training \((72\%)\), validation \((10\%)\) and testing \((18\%)\) sets. All sets are randomly sampled from both US-101 and I-80 subsets. 

Vehicle trajectories are split into segments of 8 s, where we use
\(t_h = 3\) s of track history and a \(t_h = 5\) s prediction horizon. These 8 s segments are sampled at the dataset sampling rate of 10 Hz, and then downsampled by a factor of 2 before feeding them to the LSTMs, to reduce the model complexity \cite{deo2018convolutional}.

\begin{center}
\vspace{2mm}
\begin{table}[t]
\caption{The \textit{overall} evaluation of the baselines using the RMSE of the predicted trajectories (m) at different prediction horizons (s), with (w/) and without (w/o) using the maneuver recognition module.}
\centering 
\begin{tabular}{|c| c c | c c | c c |} 
\hline
Prediction\\ Horizon (s) & \multicolumn{2}{|c|} {S-LSTM} & \multicolumn{2}{|c|} {CSP} & \multicolumn{2}{|c|} {S-GAN}\\ 

&w/ &w/o&w/ &w/o&w/ &w/o\\
\hline \hline
1 & 0.33 & 0.33 & 0.34 & 0.34 & 0.34 & 0.34\\
2 & 0.97 & 0.98 & 0.98 & 0.99 & 0.98 & 0.98\\
3 & 1.72 & 1.79 & 1.73 & 1.77 & 1.73 & 1.76\\
4 & 2.65 & 2.85 & 2.68 & 2.79 & 2.66 & 2.78\\
5 & 3.83 & 4.19 & 3.87 & 4.11 & 3.84 & 4.09\\
\hline
\end{tabular}
\label{table:mnvrs}
\end{table}
\vspace{-5mm}
\end{center}

\begin{center}
\vspace{2mm}
\begin{table*}[t]
\caption{Performance comparison of the proposed pooling mechanism against the baselines using the RMSE of the predicted trajectories (m) at different prediction horizons (s). The \textit{overall} evaluation and the \textit{keep} maneuver results are reported.}
\centering 
\resizebox{\textwidth}{!}{
\begin{tabular}{|c| c c c c c| c c c c c|} 
\hline
Prediction\\ Horizon (s) & \multicolumn{5}{|c|} {overall} & \multicolumn{5}{|c|} {keep}\\ 

&S-LSTM&CSP&S-GAN&Polar&Polar-\(V_r\)&S-LSTM&CSP&S-GAN&Polar&Polar-\(V_r\)\\
\hline \hline
1 & 0.33 & 0.34 & 0.34 & 0.34& \textbf{0.25} & 0.37 & 0.38 & 0.38 & 0.38& \textbf{0.29} \\

2 & 0.97 & 0.98 & 0.98 & 0.96 & \textbf{0.84} & 1.01 & 1.02 & 1.02 & 1.01 & \textbf{0.89}\\

3 & 1.72 & 1.73 & 1.73 & 1.71 & \textbf{1.58} & 1.72 & 1.76 & 1.75 & 1.72 & \textbf{1.62} \\ 

4 & 2.65 & 2.68 & 2.66 & 2.66 & \textbf{2.53} & 2.57 & 2.64 & 2.61 & 2.58 &  \textbf{2.51} \\ 

5 & 3.83 & 3.87 & 3.84 & 3.85 & \textbf{3.75} & \textbf{3.60} & 3.72 & 3.66 & 3.63 &  3.62 \\
\hline
\end{tabular}}
\label{table:results_keep}
\end{table*}
\vspace{-5mm}
\end{center}

\begin{center}
\vspace{2mm}
\begin{table*}[t]
\caption{ The prediction accuracy results of \textit{merge}, \textit{left} and \textit{right} maneuvers.}
\centering 
\resizebox{\textwidth}{!}{
\begin{tabular}{|c| c c c c c| c c c c c| c c c c c|} 
\hline
Prediction\\ Horizon (s) &  \multicolumn{5}{|c|} {merge} & \multicolumn{5}{|c|} {left} & \multicolumn{5}{|c|} {right}\\ 

&S-LSTM&CSP&S-GAN&Polar&Polar-\(V_r\)&S-LSTM&CSP&S-GAN&Polar&Polar-\(V_r\)&S-LSTM&CSP&S-GAN&Polar&Polar-\(V_r\)\\
\hline\hline
1 & 0.35 & 0.36 & 0.37 & 0.34 & \textbf{0.25}  & 0.54 & 0.53 & 0.54 & 0.48 & \textbf{0.37}  & 0.60 & 0.62 & 0.58 & 0.54 & \textbf{0.39}\\

2 & 1.03 & 1.06 & 1.07 & 0.99 & \textbf{0.91}  & 1.50 & 1.50 & 1.54 & 1.41 & \textbf{1.22} & 1.93 & 1.87 & 1.76 & 1.72 & \textbf{1.38} \\

3 & 1.89 & 1.94 & 1.95 & 1.82 & \textbf{1.76} & 2.70 & 2.72 & 2.77 & 2.61 & \textbf{2.38} & 3.53 & 3.46 & 3.38 & 3.28 & \textbf{2.83} \\ 

4 & 2.93 & 2.96 & 2.93 & \textbf{2.75} & 2.81 & 4.14 & 4.18 & 4.23 & 4.09 & \textbf{3.88} & 5.47 & 5.46 & 5.35 & 5.16 & \textbf{4.63}\\ 

5 & 4.19 & 4.18 & 4.08 & \textbf{3.85} &4.04  & 5.87 & 5.88 & 5.93 & 5.83 &\textbf{5.69}  & 7.59 & 7.76 & 7.64 & 7.32 & \textbf{6.74} \\
\hline
\end{tabular}}
\label{table:results}
\end{table*}
\vspace{-5mm}
\end{center}

\subsection{Implementation Details}

We extended the model in \cite{deo2018convolutional} to: (1) incorporate our benchmark pooling strategies, and (2) use multi-variate Gaussian distribution. The SUM pooling layer used in S-LSTM \cite{alahi2016social} is implemented using a two-dimensional kernel of size (4,3). The sizes of the convolutional social pooling layers of CSP are set as given in \cite{deo2018convolutional}. The MLP used in the proposed pooling module (and in S-GAN pooling \cite{gupta2018social}) has a size of 256 and is followed by a leaky-ReLU layer. The acceleration threshold in Sec.\ref{sec:mnvr_rec} is set to  \(a=0.2 m/s^2\). The spatial grid in Fig. \ref{fig:pooling} is set to a 13 × 3 size, where each column corresponds to a single lane, and the rows are separated by a distance of 15 feet which approximately equals one car length \cite{deo2018convolutional}. To emphasize the effectiveness of the proposed pooling strategy, we set the distances \(l_{lat}\) and \(l_{long}\) to the same width and length, respectively  of the spatial grid. The model is implemented using PyTorch \cite{paszke2017automatic} and trained end-to-end.

\subsection{Evaluation metric}

All results are reported in terms of the root of the mean squared error (RMSE) of the predicted trajectories with respect to the ground truth future trajectories, over the prediction horizon. Since the LSTM models generate bi-variate Gaussian distributions, the means of the Gaussian components are used for RMSE calculation.

\subsection{Baselines}
The results of the following pooling approaches are compared:
\begin{itemize}
  \item \textbf{S-LSTM}: Social Pooling \cite{alahi2016social}. 
  \item \textbf{CSP}: Convolutional Social Pooling \cite{deo2018convolutional}. 
  \item \textbf{S-GAN}: the pooling strategy used in S-GAN \cite{gupta2018social}.
  \item \textbf{Polar}: the proposed pooling strategy where trajectories are modelled by position only in the polar coordinates \(r\) and \(\phi\), and the model outputs a bi-variate Gaussian distribution.
  \item \textbf{Polar-}\(\boldsymbol{V_r}\): the full proposed pooling strategy presenting trajectories by position and radial velocity in polar coordinates (\(r, \phi, V_r\)), and the model outputs a multi-variate Gaussian distribution.
\end{itemize}

While we employ polar representation of vehicle trajectories, other baselines (S-LSTM, CSP, S-GAN) use Euclidean lateral and longitudinal positions. The same model and testing data are used to evaluate every pooling strategy. The maneuver recognition module (Sec. \ref{sec:mnvr_rec}) was incorporated in the \textbf{Polar-}\(\boldsymbol{V_r}\) model only. At the evaluation time, this model outputs the maximum \textit{a posteriori} probability (MAP) trajectory estimate corresponding to the maneuver classes having the maximum probability.

\subsection{Results and Discussion}

Besides reporting the \textit{overall} prediction accuracy, we evaluate the baselines on the following maneuvers: (1) car-following by keeping the same lane (\textit{keep}), (2) merging from the onramp lane (\textit{merge}), (3) left lane-change (\textit{left}), and (4) right lane-change (\textit{right}). The \textit{overall} performance of the baseline approaches in Table \ref{table:mnvrs} implies two findings\footnote{We also noted the same findings for the maneuver-based evaluation.}: the baselines have comparable prediction accuracy, and RMSE accuracy was not improved by using the maneuver recognition module. While the former motivates for a different direction of improvement (other than the pooling layer type and the neighborhood structure), the latter suggests feeding higher-order information to the pooling operation.   

RMSE results of the maneuver-based evaluation are reported in Tables \ref{table:results_keep} (\textit{overall} and \textit{keep}) and \ref{table:results} (\textit{merge}, \textit{left} and \textit{right}). We note that the overall prediction accuracy is dominated by the keep maneuver accuracy. This is due to the inherent biasness in the data: as drivers tend to keep their lanes most of the time, the dataset provides fewer lane-change examples. However, Our proposed solution tackled this problem without changing the dataset (e.g. by augmentation). 

Representing trajectories using relative distance and angle (\textbf{Polar}) is shown to outperform the Euclidean representation adopted by the state-of-the-art strategies, especially at the lane-change and merging maneuvers. The main reason of this improvement is employing the vehicle orientation which makes the pooling operation more maneuver-aware and hence the learning process more efficient.   

The maneuver-recognition module estimates the location and acceleration maneuver classes. The input to this module is the embedding of the vehicle dynamics encoded by the LSTM hidden states. It is reasonable to expect that a single-layer LSTM can encode the change in vehicle location into states that are informative for estimating the change in location. However, the accuracy of such networks degrades when encoding the higher-order dynamics necessary for estimating the acceleration. Therefore, we augment the trajectory representation by the vehicle velocity. More importantly and inline with our proposed polar representation, we employ radial motion velocity that encodes vehicle orientation. This defines our full proposed model \textbf{Polar-}\(\boldsymbol{V_r}\) that further improved the prediction accuracy.

\section{CONCLUSIONS}

We introduced a novel pooling mechanism to improve the accuracy of predicting vehicle behaviors when performing lane change and highway merging maneuvers. To be able to capture a maneuver-aware interaction, our pooling strategy employs the orientation and radial motion velocity of the interacting vehicles. We demonstrated how incorporating such higher-order information leads to a more efficient learning process. Our maneuver-aware pooling mechanism outperforms the state-of-the-art pooling approaches in predicting vehicle trajectories in both the overall and maneuver-based evaluation. 

One limitation of this work is predicting the future of a single agent based on pooling the states of its neighbors. This can be improved by jointly predicting the future maneuvers and trajectories of all the agents involved in a scene. Extending the current model to perform such joint estimation will be addressed in our future work.

\addtolength{\textheight}{-12cm}   




\section*{ACKNOWLEDGMENT}

The work described in this paper is supported by VeriCAV project, which is funded by the Centre for Connected and Autonomous Vehicles, via Innovate UK (Grant number 104527). This work was undertaken on ARC4, part of the High Performance Computing facilities at the University of Leeds, UK.

%

\bibliographystyle{IEEEtran}
\bibliography{IEEEabrv,pooling_ref.bib}

\end{document}